# Predicting Terrain Mechanical Properties in Sight for Planetary Rovers with Semantic Clues

Ruyi Zhou, Wenhao Feng, Huaiguang Yang*, Haibo Gao, Nan Li, Zongquan Deng, and Liang Ding*, *Senior Member, IEEE*

***Abstract*—Non-geometric mobility hazards such as rover slippage and sinkage posing great challenges to costly planetary missions are closely related to the mechanical properties of terrain. In-situ proprioceptive processes for rovers to estimate terrain mechanical properties need to experience different slip as well as sinkage and are helpless to untraversed regions. This paper proposes to predict terrain mechanical properties with vision in the distance, which expands the sensing range to the whole view and can partly halt potential slippage and sinkage hazards in the planning stage. A semantic-based method is designed to predict bearing and shearing properties of terrain in two stages connected with semantic clues. The former segmentation phase segments terrain with a light-weighted network promising to be applied onboard with competitive 93% accuracy and high recall rate over 96%, while the latter inference phase predicts terrain properties in a quantitative manner based on human-like inference principles. The prediction results in several test routes are 12.5% and 10.8% in full-scale error and help to plan appropriate strategies to avoid suffering non-geometric hazards.**

***Index Terms*—terrain sensing system, wheeled robots, terrain semantic segmentation**

## I. Introduction

OVER the past half century, exploring planets such as Mars and the Moon with rovers has brought many valuable scientific discoveries [1-3] and boosted the growth of extraterrestrial geology. During the exploration, rovers acting as scientists navigate around, collect data as well as perform science experiments to analyze the minerals or elements of rocks and soil. Even if there are many approaches and tools [4,5] developed to improve autonomy during planetary exploration, the traverses on rough and deformable terrain is still highly artificial [6]. One of the crucial factors hampers fully autonomous traverses lies in non-geometric hazards such as deformable ripples and slippery sand dune that are related to the bearing and shearing properties of terrain to a great extent. Terrain of these characteristics is prone to plunge the rover into danger or impede the pace of mission with excessive normal sinkage or tangential slip. Accidents have already happened on Spirit [7] and again on Opportunity [8].

Researchers has presented some methods to identify mechanical characteristics of terrain for rovers, most of which are based on in-situ measurements. For instance, slip and sinkage behavior of the rover yield data to identify properties of terrain indirectly with semi-empirical terramechanics models [9,10]. However, these interaction-based methods are limited to traversed terrains but helpless to untraversed regions. Once the rover drives into hazardous area, it may pay time and effort to painstaking struggling or even too late to escape from entrapment. More accurate and direct methods performing penetration test and shear test with a bevameter [11,12] for characterization add additional scientific payloads to the rover and yet don't get rid of range limitation.

In contrast, vision enjoys a wide perceptive field as well as abundant information. It plays a core role in planetary autonomous navigation through scanning surroundings and avoiding potential hazards before getting access to. If intelligent rovers are capable of distinguishing terrains and inferring their mechanical properties at a distance, a number of non-geometric hazards can be halted in the planning stage, reducing risks of being involved in danger. In the meanwhile, a big picture review of the surroundings ahead can spare time for rovers to switch adaptive locomotion mode before dashing into challenge but scientific-valuable terrains. However, how to predict mechanical properties of terrain with visual data remains a tough problem.

The ability to predict mechanical properties of the scene is quite common among high-intelligent creature. Humans can easily identify and categorize materials at a glance, and predict their properties based on common sense built up over a lifetime of experience. Such perception ability using data from visual modality to infer feelings of haptic is not only revealed in humans but also in great apes with evidence [13,14]. Moreover, in cognitive science, the function of language is deemed as an integrative mediating process offering a well-formed semantic structure for reasoning between different sensing modalities [15].

Inspired of these points in cognitive science, we equip the rover with a similar ability of inferring mechanical properties of the terrain based on vision. A semantic-guided prediction

This work was supported in part by the National Natural Science Foundation of China under Grant 51822502, Grant 91948202 and Grant 52005122, in part by the National Key Research and Development Program Funds for the Central Universities under Grant HIT.BRETIV.201903, and in part by the "111 Project" under Grant B07018. *(Corresponding authors: Huaiguang Yang; Liang Ding.)*

Ruyi Zhou, Wenhao Feng, Huaiguang Yang, Haibo Gao, Nan Li, Zongquan Deng and Liang Ding are with the State Key Laboratory of Robotics and Systems, Harbin Institute of Technology, Harbin 150080, China (e-mail: zhouryhit@gmail.com; 1160800125@stu.hit.edu.cn, yanghuaiguang_hit@163.com, gaohaibo@hit.edu.cn, lnlinanln@126.com, dengzq@hit.edu.cn, liangding@hit.edu.cn).

method for terrain bearing and shearing properties is proposed to realize more accurate perception of the surrounding environment for rovers in a quantitative manner. The main contributions lie in the following three aspects:

1) A semantic-guided terrain mechanical property prediction method is proposed to make dense prediction of the untraversed terrain in terms of normal bearing and tangential shearing properties in a quantitative manner and it enjoys a wide perceptive range as well as fine-grained results for autonomous non-geometric hazards alarm.

2) A light-weighted terrain semantic segmentation model is achieved with satisfactory accuracy and speed considering the limited onboard calculation resources on rovers.

3) A well-annotated semantic segmentation dataset containing multiple terrains in an emulated Martian environment is provided as a platform for performance evaluation of planetary scene understanding.

This paper is organized as follows: Section II describes and analyzes related work about interaction-based and visual-based terrain perception methods in terms of mechanical properties. Section III introduces the methodology of semantic-guided terrain mechanical properties prediction in a rough overview with three detailed parts. Section IV presents data collection and dataset setup. Section V discusses experiments and results of terrain semantic segmentation and terrain mechanical properties estimation. Section VI summarizes the main conclusions of this research and suggests guidelines for future work.

## II. Related Work

Most methods in terrain mechanical properties perception are based on force, torque or vibration. Key parameters charactering terrain shearing strength, cohesion and internal friction angle, are estimated on line when a rover moving on loose sand [9]. A more accurate step-by-step identification of all 8 unknown soil parameters with closed form analytical equation is conducted with different slip states [16]. Terrain parameters composed of 4 characteristic parameters and 72 identification parameters are also estimated with simulant regolith of three states (loose, natural, and compact states) based on different hardness [17]. These proprioception methods are limited in range. Rovers may already be entrapped in soft and slippery area and too late to make large changes. In addition, most of these interaction-based methods are only applicable on soft soil pursuing high estimation accuracy and ease-of-use but not universal to more types of terrains. Attempts of proprioception on various terrains appeared on terrain classification [18]. Replacing vibration of wheel-terrain interaction with state information collected during leg-terrain interaction, similar ideas are also applicable to legged robots of different scales [19,20]. However, they are confined to classification, a qualitative result, which makes it difficult to measure the traversability and safety margin for autonomous navigation and not break free from range limitation.

Vision is the core of perception which takes up of 83% of all information obtained by humans [21]. The same principle is almost also identical to the planetary perception system. Therefore, there are numerous studies devoted to understanding the environment through vision. The most common one is visual-based terrain classification on patch-level [22] or pixel-level [23]. A Deep Encoding Pooling Network (DEP) is established integrating orderless texture details and local spatial information for ground terrain recognition in outdoor environment [24]. Admittedly, there are dazzled methods to classify or segment artificial scenario containing pedestrian and vehicles in a supervised manner for autonomous driving [25], but applications on planetary environment are still research-strapped for short of public well-annotated dataset. Another possible is to learn with the assistant of an interaction-based classifier generating corresponding labels and enable rovers to infer far from near [26]. With the involvement of more diverse methods like co- and self-training [27], deep learning [28] and multi-sensor fusion [29], and improvement on accuracy as well as less data demand, solutions still not get rid of qualitative characterization. The classification results can only provide fuzzy criteria for traversability judgment which put the rover into a dilemma when its routes head for targets of scientific value besieged by challenging terrains of different complexity. In such cases, the rover needs quantitative results in terms of terrain properties for more adequate assessment.

One of the earliest examples of predicting specific terrain properties in a quantitative way for planetary applications is present on slip [30,31] which, to be more exact, is a state parameter reflecting wheel-terrain interaction characteristics rather than intrinsic terrain properties. Accordingly, it is not compatible among different rovers as involved in a slip regression model greatly related to rover configuration such as lugs and size. Similar research committed to build more accurate slope-slip characteristic curve for prediction with more experiments on different terrains [30,32]. More recent demonstrations include terrain characteristics prediction with ANYmal robot [33] and Athena rover [34]. The former generates dense predictions of terrain attributes which are not specific terrain parameters but a combined ground-reaction score relevant to ground reaction forces during the stance phase. The latter estimates terrain parameters of each image patches using a simplified terramechanics model for driving energy estimation.

In contrast, the proposed semantic-guided terrain mechanical property inference method predicts dense feature map of intrinsic and specific terrain bearing and shearing characteristics for the whole visual perceptive field. The whole work is partly about terrain classification on the pixel-level but goes beyond semantics and outputs quantitative prediction results of terrain mechanical properties.

## III. Semantics-Guided Terrain Mechanical Properties Inference

### A. Framework

The semantic-guided terrain mechanical properties inference does not establish a direct mapping from the appearance to properties end to end but instead invokes the associations of semantic consistency which implies properties in nature. An overview of the terrain mechanical properties prediction method consisting of two phases linked with semantic clues is shown in Fig. 1. The segmentation phase gets probabilistic semantic prediction results for each pixel with a convolutional neural network. The inference phase transforms the results from semantic feature space to continuous mechanical property feature space based on empirical model built up over traverse experience.

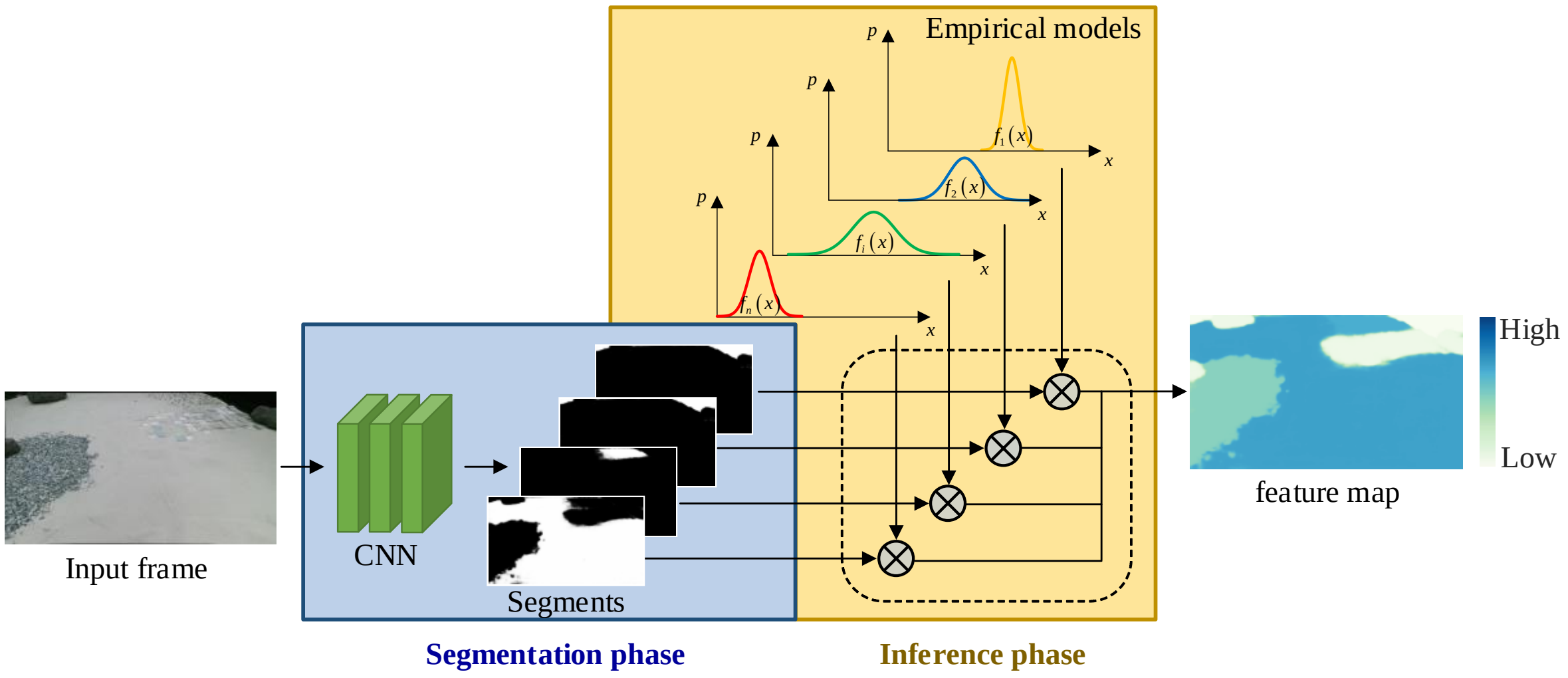

Fig. 1. Pipeline of semantics-guided terrain mechanical properties inference.

## B. *Light-weighted Terrain Semantic Segmentation*

Since the computing and storage resources on a rover is desperately limited, the lighter the terrain semantic segmentation model is, the more suitable it is to be deployed onboard.

To simplify the model structure and reduce model parameters, considerable universal methods or modules have been come up with, mainly including (1) convolutional separation: a standard convolutional operation is decomposed into multiple steps to reduce involved hyper-parameters by spatial separable convolution [35] or depthwise separable convolution [36]; (2) network compression: eliminate redundancy in network connections or channels by hashing [37], pruning [38], vector quantization [39], and shrinking [40]; (3) dilated convolution [41]: zeros are inserted between each element in the convolutional kernel to expand the effective receptive field while without increase parameters.

Taking a page from the successful practice of lightweight neural network on mobile devices, we select the Depth-wise Asymmetric Bottleneck Network (DABNet) [42], a lightweight model with fast inference speed and competitive accuracy, as a basic structure for our tasks. The DABNet (see Fig. 2) learns robust local and contextual features jointly with Depthwise Asymmetric Bottleneck (DAB) module which adopts depth-wise asymmetric convolution and dilated convolution. Two DAB blocks containing several consecutive DAB modules for dense feature extraction is elaborately designed. The first and the second DAB block consist of 3 and 6 DAB modules in the original DABNet, respectively.

Considering the peculiarity of terrain, the DABNet is further simplified. The original network is designed oriented toward indoor or autonomous driving scenes where targets are almost about objects with particular shapes. As opposed to objects, most terrains are generally not in specific shapes except for rocks, nor are they closed in regular boundaries. Therefore, their attributes are indeed locally recognizable [43] and terrain recognition depends more on texture features usually extracted in the shallow layers rather than context information in deeper layers. With this in mind, the number of DAB modules is further decreased to 2 and 2 in the first and second DAB blocks, and we named it DABNet-Lite. The number of DAB modules in each DAB Block is explored in Section V.

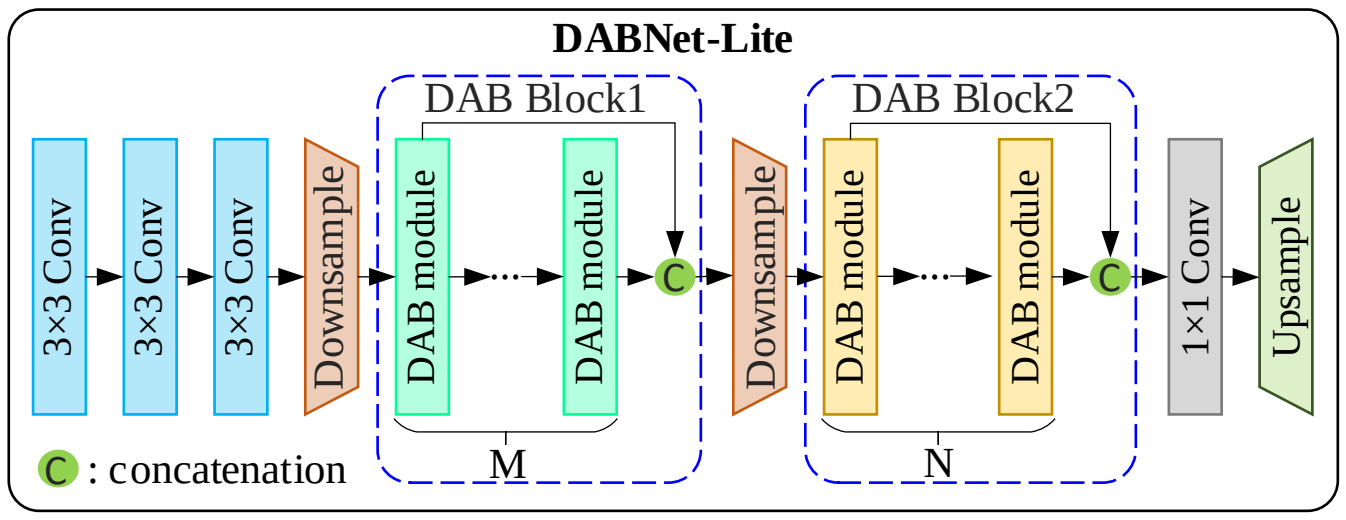

Fig. 2. The detailed semantic segmentation phase.

Considering data imbalance of various terrain categories in training set, the weighted softmax with cross-entropy loss is employed as the optimization goal for this pixel-wise classification task:

$$L_{\mathrm{WCE}} = -\frac{1}{W \times H} \sum_{x=1}^{W} \sum_{y=1}^{H} \sum_{k=1}^{K} \beta_k p_{x,y}^{k} \log\left(\hat{p}_{x,y}^{k}\right) \tag{1}$$

$$\hat{p}_{x,y}^{k} = \frac{\exp\left(\alpha_{x,y}^{k}\right)}{\sum_k \exp\left(\alpha_{x,y}^{k}\right)} \tag{2}$$

Where $\beta_k$ is the weight of the $k$-th terrain category, $p_{x,y}^{k}$ is the ground truth of the pixel in $x$ row, $y$ line to the $k$-th category in one-hot code, $\hat{p}_{x,y}^{k}$ is the softmax of $\alpha_{x,y}^{k}$ which is the predicted probability of the pixel in $x$ row, $y$ line belonging to the $k$-th category. The weight $\beta_k$ is calculated as

$$\beta_k = \frac{\ln\left(p_k + c\right)^{-1}}{\sum_{k=1}^{K} \ln\left(p_k + c\right)^{-1}} \tag{3}$$

where $p_k$ is the proportion of the $k$-th terrain category in the training set and $c$ is a constant, set in 1.1 in the experiment.

Some data augmentation measures are also taken to diversify training samples. Random scale, random horizontal flip and brightness adjustment are employed on images of the training set. The random scale operation resizes color images with a scale between 0.5 and 2. If the scale factor is larger than 1, the scaled images are randomly cropped to fixed size for training, and if it is less than 1, the scaled images are filled with default values on the rim. Meanwhile the brightness of color images is also scaled between 0.9 and 1.1 randomly to cope with conditions in different brightness.

### *C. Regression of Terrain Mechanical Properties Model*

Along the way of driving, the rover's states are mainly involved in wheel-terrain interaction shown in Fig. 3. From the perspective of mechanical analysis, the wheel-terrain interaction can be decoupled into the pressure-bearing effect along the normal direction and the shearing effect along the tangential direction. Therefore, the main mechanical properties of the terrain can be characterized with the normal bearing characteristics and the tangential shearing characteristics [44].

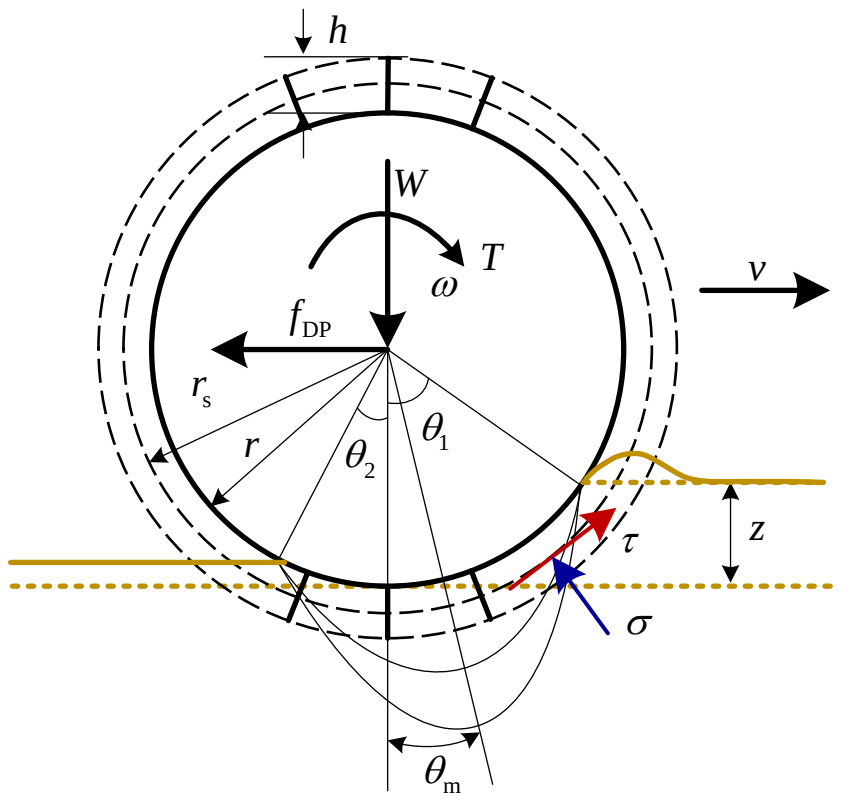

Fig. 3. The wheel-terrain interaction model. $\theta_1$ is the entrance angle at which the wheel begins to contact the soil; $F_N$ is the normal force; $M_R$ is the resistance provided by forward movement; $s$ is the slip ratio; $\theta_1'$ is the equivalent entrance angle considering lug effect; $r$ is the wheel radius; $b$ is the width of the wheel; $r_s$ is the equivalent shearing radius.

In conventional terramechanical models [45,46], there are two groups of parameters, $\{k_c, k_\varphi, N\}$ and $\{c, K, \varphi\}$, characterizing the bearing and shearing characteristics while playing different roles according to their sensitivities, as shown in Table I. The sinkage exponent $N$ and internal friction angle $\varphi$ play dominant role in these two groups [47], reflecting the trend of overall changes on properties with significant fluctuation, while the rest parameters work to fit the characteristics subtly. Consequently, dominant parameters in both groups are used to represent bearing and shearing properties for simplified characterization instead of six parameters when non-dominant parameters are fixed in typical value. In this case, sinkage exponent $N$ is regarded as the stiffness parameter of equivalent bearing property. The larger the equivalent bearing property is, the stiffer the terrain surface is. In the meantime, the internal friction angle $\varphi$ is viewed as the friction parameter of equivalent shearing property. The larger the equivalent shearing property is, the more drawbar pull terrain surface can offer to move rovers forward.

TABLE I
COMPREHENSIVE RESULTS OF MODELS IN DIFFERENT CONFIGURATIONS

| Characteristics | Parameter group | Meaning | Dominant parameter |
|---|---|---|---|
| Bearing | $\{k_c, k_\varphi, N\}$ | $k_c$: cohesive modulus of the soil<br>$k_\varphi$: frictional modulus<br>$N$: sinkage exponent | $N$ |
| Shearing | $\{c, K, \varphi\}$ | $c$: cohesion of the soil<br>$K$: shearing deformation modulus<br>$\varphi$: internal friction angle | $\varphi$ |

It is assumed that the statistical characteristics of the bearing and shearing properties of terrain are normally distributed. Then, the probability density function (PDF) of the $i$-th category of terrain with respect to property $x$ can be represented in $f_i(x)$ with mean and variance. The maximum likelihood estimation method (MLE) is used to estimate the mean and variance based on identification results, and the calculation formula is shown below:

$$\mu = \frac{1}{n}\sum_{i=1}^{n} x_i \tag{4}$$

$$\sigma^2 = \frac{1}{n}\sum_{i=1}^{n}\left(\mu - x_i\right)^2 \tag{5}$$

where, $\mu$ and $\sigma^2$ represent the mean and variance of the property, respectively. $x$ is the parameter of the property, including sinkage exponent $N$ and internal fiction angle $\varphi$ in this study. $n$ is the number of samples involved in estimation.

To model the distribution of mechanical properties over different terrains, dominant parameters of terrain characteristics are estimated based on an analytical dominant parameter estimation model with collected interactive data. The analytical estimation model of the dominant parameters [47] is deduced in our earlier work as follows:

$$\begin{cases} N = \dfrac{\ln f_1 - \ln\left[4r\theta_1 r_s\left(k_c + k_\varphi b\right)\cos\dfrac{\theta_1}{2}\left(1-\cos\dfrac{\theta_1}{2}\right)\right]}{\ln\left[r\left(\cos\dfrac{\theta_1}{2} - \cos\theta_1\right)\right]} \\ \varphi = \arctan\left\{-2r\sin\dfrac{\theta_1}{2}\sin\theta_1\left[bcr_s^2\theta_1 f_2 - 2M_R\exp\left[r_s\left(s\sin\theta_1' + \sin\dfrac{\theta_1}{2} + \theta_1'\right)\Big/K\right]\right]\Big/\left[r_s f_1 f_2\left(\cos\dfrac{\theta_1}{2}+1\right)\right]\right\} \end{cases} \tag{6}$$

where $f_1 = F_N\theta_1^2 r_s + 4M_R\sin\theta_1 - 8M_R\sin\dfrac{\theta_1}{2}$,

$$f_2 = \exp[r_s(s\sin\theta_1' + \sin\frac{\theta_1}{2} + \theta')/K] - \exp[r_s(2s\sin\frac{\theta_1}{2} + s\sin\theta_1' + \theta_1)/2K].$$

In this model, the analytical expression of sinkage exponent $N$ is a function of $\{\theta_1, F_N, M_R, k_c, k_\varphi, r, b, r_s\}$, and that of the internal friction angle $\varphi$ is a function of $\{s, \theta_1, \theta_1', F_N, M_R, c, K, r, b, r_s\}$. Each group of variables consists of three parts of parameters, including system state parameters (part I), size parameters of rover wheels (part II) and non-dominant terrain mechanical parameters (part III), as listed in Table II. The former two parts of parameters can be collected, hence the dominant parameters can be estimated and regarded as equivalent parameters reflecting overall bearing as well as shearing properties of terrain when non-dominant parameters are fixed with empirical values.

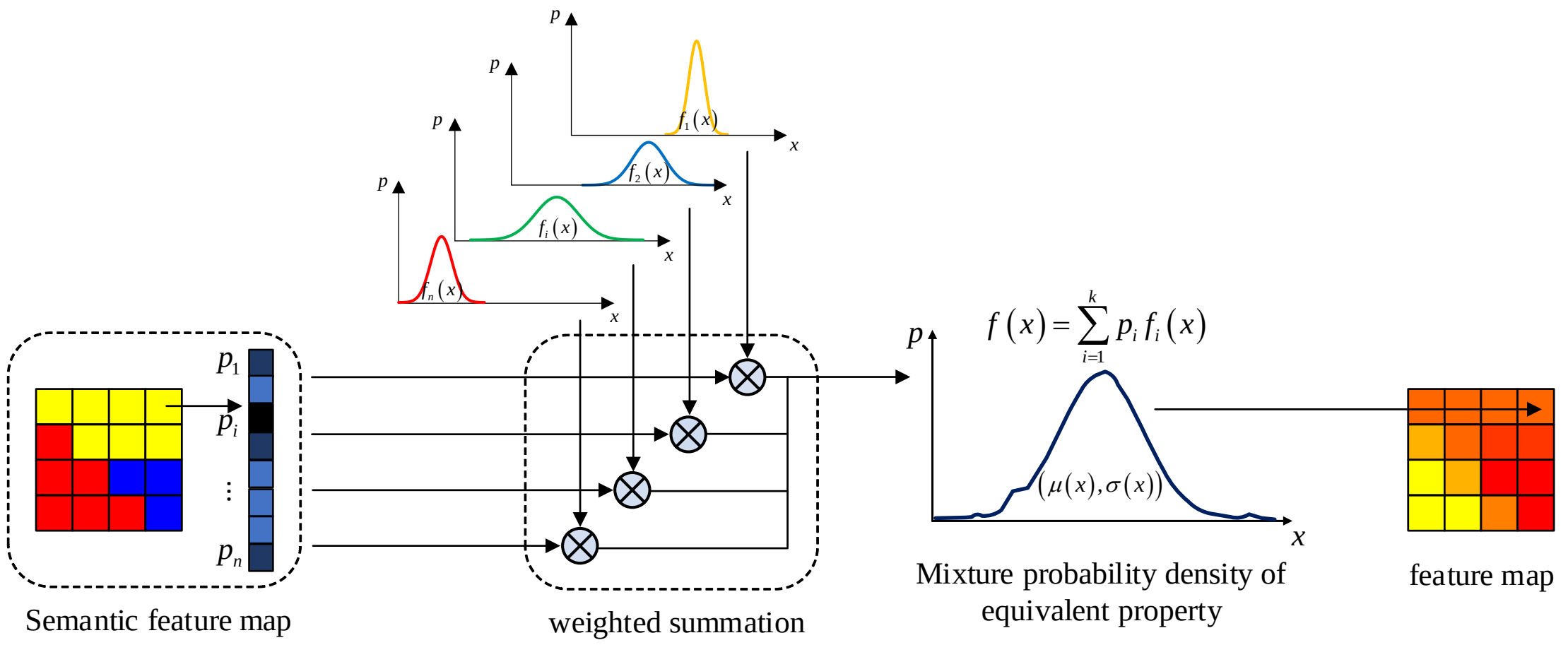

Fig. 4. Pipeline of semantics-guided terrain mechanical properties inference.

TABLE II
VARIABLES IN ANALYTICAL MODEL OF BEARING AND SHEARING CHARACTERISTICS

| Characteristics | I | II | III |
|---|---|---|---|
| Bearing | $\{\theta_1, F_N, M_R\}$ | $\{r, b, r_s\}$ | $\{k_c, k_\varphi\}$ |
| Shearing | $\{s, \theta_1, \theta_1', F_N, M_R\}$ | | $\{c, K\}$ |

The collected data provides a series of measured force, torque and motion status of each wheel on the rover with timestamps. They are down-sampled to the same frequency and used to identify the equivalent bearing and shearing property. Then, the identification results are grouped according to the terrain category corresponding to each piece of data for the probability density model regression of each terrain with corresponding mechanical properties.

### *D. Semantics-Guided Inference Principles*

Based on the regression model of mechanical properties for each terrain category, each pixel with semantic label is transformed into mechanical property domain as illustrated in Fig. 4. Here, semantics are taken as clues to build the connection between the terrain categories and corresponding mechanical properties. To illustrate how to bridge these two domains, a pitch of semantic feature map is taken as an example. Assume there are $k$ categories of terrain, for each pixel in semantic feature map, the $i$-th category of terrain is predicted with probability $p_i$ after segmentation phase. The probability values satisfy $\sum_{i=1}^{k} p_i = 1$. The probability of each terrain serves as the weight to predict terrain mechanical parameters. Then, the equivalent mechanical characteristic parameter $x$ based on semantic segmentation results can be congregated through mixture principles of probability density distribution. The mixture distribution is represented as

$$f(x) = \sum_{i=1}^{k} p_i f_i(x). \tag{7}$$

Meanwhile, in the PDF of various terrain mechanical parameters, the mean value and variance of the $i$-th terrain category are expressed as $\mu_i$ and $\sigma_i$, $(1 \le i \le k)$. Furthermore, the predicted mean and variance of the equivalent mechanical characteristic parameter can be expressed by the following formula:

$$\mu(x) = \sum_{i=1}^{k} p_i \int_{-\infty}^{+\infty} x f_i(x)\mathrm{d}x = \sum_{i=1}^{k} p_i \mu_i \tag{8}$$

$$\sigma^2(x) = \sum_{i=1}^{k} p_i \int_{-\infty}^{+\infty} (x-\mu)^2 f_i(x)\mathrm{d}x = \sum_{i=1}^{k} p_i \left(\mu_i^2 + \sigma_i^2\right) - \mu_i^2 \tag{9}$$

where $\mu(x)$ and $\sigma^2(x)$ are the mean and variance of the equivalent mechanical characteristic parameter $x$ of terrain.

According to the above prediction formula, the terrain bearing and shearing properties of the full field of view are predicted based on the output feature map generated from semantic segmentation phase. The standard deviations of the predicted results serve as confidence intervals, which lay the foundation for path planning considering terrain mechanical properties.

## IV. DATA COLLECTION AND DATASET SETUP

### *A. Experimental Setup*

A three-wheeled planetary rover prototype embodying two driving wheels and a follower was used to collect images and wheel-terrain interaction data in an emulated Martian test yard as shown in Fig 5. Wheels of rover are evenly equipped with lugs on the cylindrical surface to strengthen locomotion performance on rugged or deformable terrain, and their specifications are illustrated in Table II. In the inside, F/T sensors and encoders are mounted in-line with wheels and calibrated before experiments. On the mast of the rover, an RGB-D sensor is installed for navigation, capturing color images and making dense depth measurements. It is rotatable in pitch angle which is modulated before trials to get multi-view data but locked on a fixed base during rover traversing. Fig. 6 shows various terrain categories randomly scattered in the test yard to emulate realistic Martian environment, including sand, stony sand, gravel, bedrock and rock. The scenario is rearranged after a few trials of different traverses to diversify data. Besides, the small yard is also surrounded with a motion capture system hanging from the ceiling to collect accurate postures of the rover in movement.

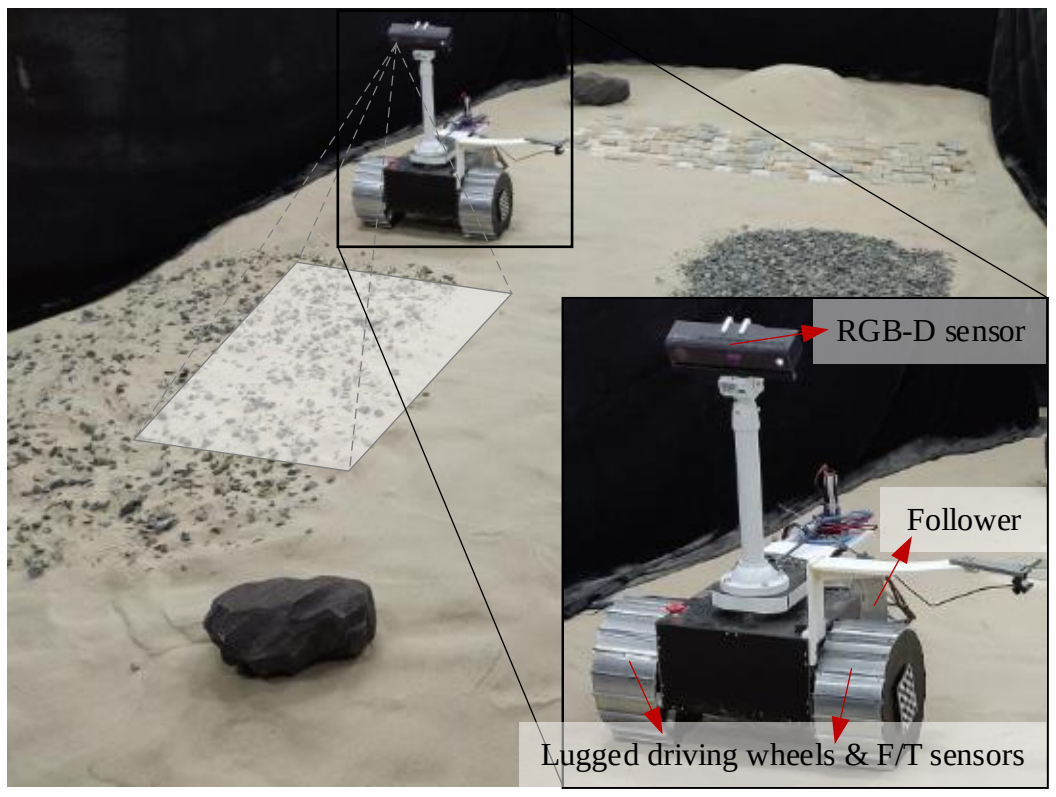

Fig. 5. Experimental environment and rover prototype.

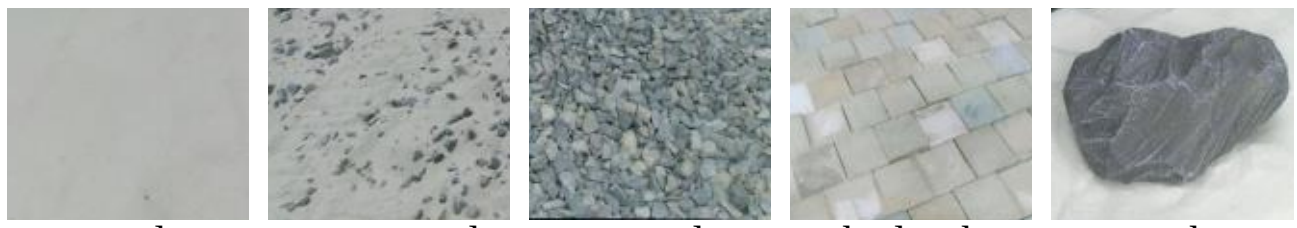

sand stony sand gravel bedrock rock

Fig. 6. Various terrain types in experimental environment.

## B. Data Collection

Data has been collected during the rover travels around the emulated Martian environment simulating planetary exploration with a probe. The wheel's vertical load, motor driving torque are measured by six-axis F/T sensors at 100Hz and the angular velocity of each wheel is recorded with an encoder at 100Hz. The motion capture system captures the postures of the rover at 120Hz. The RGB-D sensor collects color images in 540×960 resolution and depth images. The frequency of both of them are 5Hz. All of the data collected are arranged with timestamps.

In order to identify dominant parameters of terrain mechanical properties, the wheel sinkage $z$ is calculated indirectly with the depth data and rover postures. In addition, the slip ratio $s$ of wheels is jointly solved based on the actual vehicle velocity $v$ derived from positions and the angular velocity $\omega$ of the wheel by the following formula:

$$s = (r_s\omega - v)/(r_s\omega), \quad (0 \le s \le 1). \tag{10}$$

The typical values of the non-dominant parameters used in parameter identification experiments are also included in Table III.

TABLE III
PARAMETERS SPECIFICATION

| Parameters | Value | Parameters | Value |
|---|---|---|---|
| $b$ (mm) | 150 | $k_c$ (kPa/m$^{N-1}$) | 100 |
| $r$ (mm) | 140 | $k_\varphi$ (kPa/m$^N$) | 1400 |
| $h$ (mm) | 10 | $c$ (kPa) | 1 |
| $r_s$ (mm) | 145 | $K$ (mm) | 16 |

Since the rover is not equipped with active suspension, climbing a towering rock will damage the wheel structure irreversibly. Therefore, the interaction data is only collected when driving on bedrock, sand, gravel and stony sand. Data collection was conducted twice in the spring and fall of 2019, and 26 and 24 trials were carried out respectively to get as diverse data as possible.

## C. Semantic Segmentation Dataset

The terrain images collected are manually annotated and arranged into a semantic segmentation dataset. In order to improve annotation efficiency, image labels have been mainly obtained in two annotation methods.

### 1) Semi-automatic Full Labelling

Since the images are continuously recorded at a fixed frequency when the rover is traveling, adjacent images collected have certain overlapping area. Furthermore, the depth information and rover postures are also available in our experiment. Taking advantage of these convenience, a semi-automatic labeling method was developed in the interest of improving the labeling efficiency. For images collected in a consecutive path, the labelled markers on few detailed annotated images can be projected to 3D space with assistant of depth images and rover's body postures, then re-projected to unlabeled image plane though transformation matrix deduced from rover's postures. It gets annotation for every image when only a few images are detailed annotated artificially. The semi-automatic labeling method is also applicable to annotate data with spatial continuity is a static scene such as semantic segmentation annotation or instance annotation in the wild scene.

### 2) Random Partial Labelling

Since the data volume in the dataset is not big and the dependency on context information for segmentation of terrains is weaker than object-oriented counterpart, to learn more robust terrain features and reduce the possibility of overfitting, labelling of random shapes in some definitive areas is carried out instead of carefully annotation of the whole image. To prevent specific shape features are learned during training with these samples, the shapes of these partial annotations are in diversity, as shown in Fig. 7. In addition, only the area marked gets involved in loss calculation.

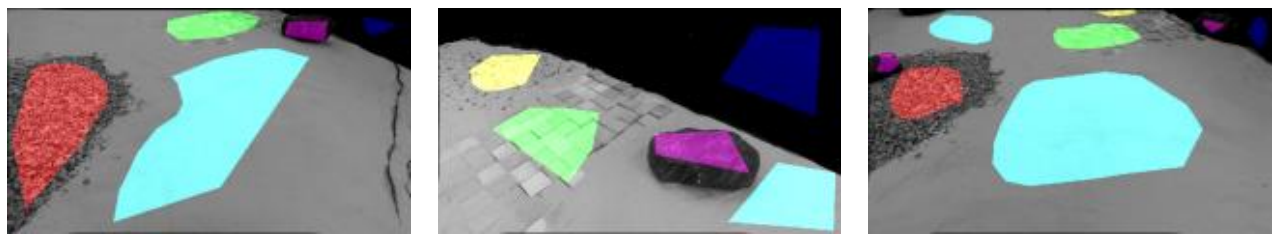

Fig. 7. Examples of random partial labeling.

In order to reduce the similarity of the image, downsampling is carried out and not all of the collected images are used in the dataset. The well-annotated dataset is named **Common Terrain in Emulated Mars** (CTEM) containing 2375 samples of 540×960 resolution. Among them, 500 images of 5 trials are partially labeld for training. The remaining images are aritificially annotated or generated with semi-automatic labelling method. All of the samples are divided into training set, validation set and test set according to different trials to avoid similarity, where the number of images are 1647, 320 and 408 seperately. The CTEM dataset will be available online in the near future.

# V. Experiments and Results

## A. Results of Terrain Semantic Segmentation

The DABNet-Lite is implemented and trained on the Pytorch platform [48] with one GTX 1080 GPU, CUDA 9.0 and cuDNN V7. Optimization is carried out using Adam [49]. with batch-size of 8, momentum of 0.9 and weight decay of 5e$^{-4}$ in training. The initial learning rate is set to 1e$^{-3}$ and exponential decay strategy is adopted with a decay rate of 0.96.

After epochs of training, the optimization objective and average accuracy on the training and validation set converged and reached a stable state. The mean interest of union (mIOU) of the trained model in the test set was 93.0%, and the statistical results of the predictions on the test set are demonstrated in Fig. 8. As we can see in the confusion matrix, each kind of terrain has a high recall rate over 96%. Stony soil and bedrock are in part predicted as soil as shown, probably because the soil occupies the largest proportion among all terrains and they are distinguishable at the boundary, easily resulting in misjudgment.

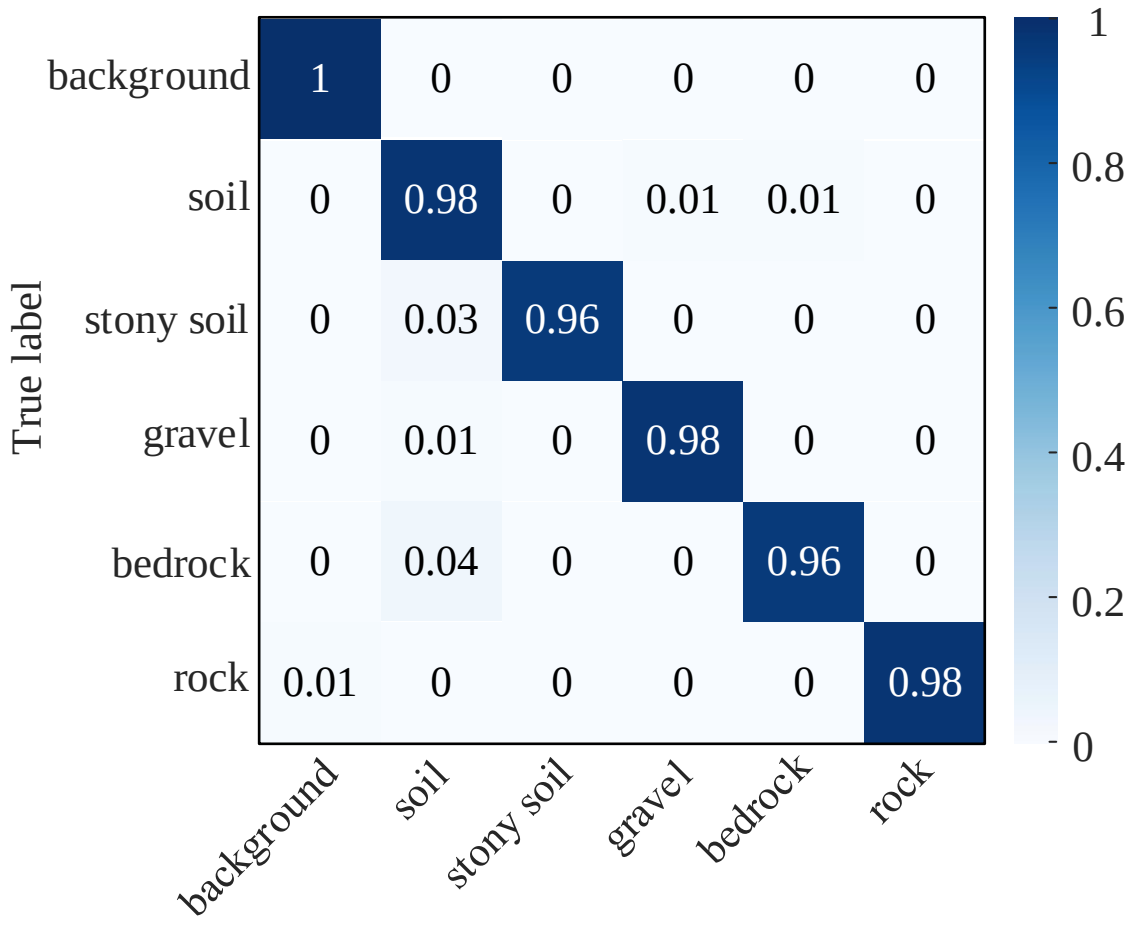

Fig. 8. Confusion matrix of terrain semantic segmentation.

Fig. 9 shows the contrast between the prediction and the groud truth in segmentation. Overall, the prediction results of the trained model are in great agreement with the labeled results, and basically reached a satisfactory level. In careful comparison, undesirable prediction appears around the boundaries that supposed to be sharp but turning out smooth. It is almost due to the bilinear interpolation with 8 times magnification in the last layer of the DABNet and some details of edges are lost during such a sharp magnification process.

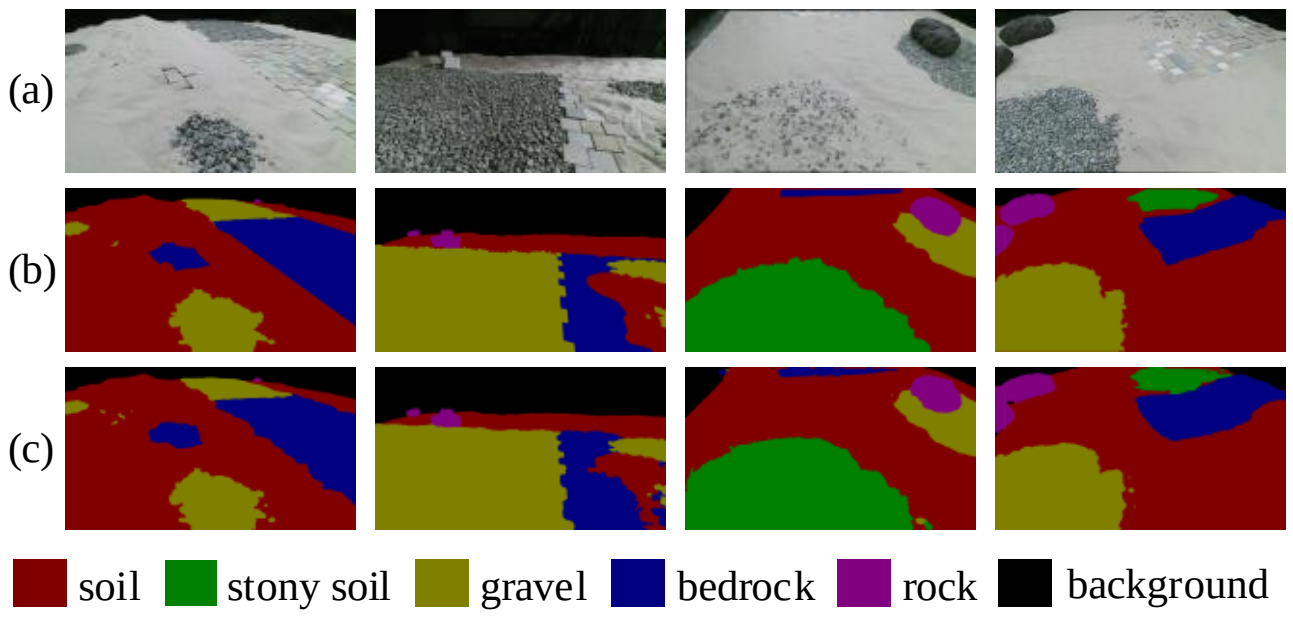

Fig. 9. Visualization of terrain segmentation results, (a) raw images, (b) the ground truth, (c) prediction.

### 1) *Architecture Evaluation*

In general, the performance of a network with more layers is better than that of shallow structure, but the corresponding real-time performance will decline with the increase of computing cost and memory consumption. To make a tradeoff between accuracy and speed, models with different numbers of DAB modules in two DAB blocks under the aforementioned architecture are explored. The two aspects of performance for different models are demonstrated in Table IV. It is presented that the mIOU of network with less DAB modules is slightly decreased when trained in the same batch size, but its number of parameters is reduced to nearly a half. The float operations (FLOPs) and prediction frames per second (FPS) are significantly improved demonstrating the prediction speed increase by 47%. When the batch size of training is doubled, mIOU rebounded to a certain extent, making up for the loss of training accuracy. Considering both accuracy and speed, the number of DAB modules in two DAB Blocks are decreased to 2 and 2, which is selected as a relative optimal structure and used in the terrain mechanical prediction experiments. Moreover, the dilation rate in DAB block1 and block2 are 2 and 4 respectively.

TABLE IV
COMPREHENSIVE RESULTS OF MODELS IN DIFFERENT CONFIGURATIONS

| M | N | Batch-size | mIOU(%) | Parameters(M) | FLOPs(G) | FPS |
|---|---|---|---|---|---|---|
| 3 | 6 | 4 | **93.2** | 0.75 | 10.4 | 78.9 |
| 2 | 4 | 4 | 92.9 | 0.56 | 8.3 | 103.5 |
| 2 | 2 | 4 | 92.7 | **0.39** | **6.9** | **116.1** |
| 2 | 2 | 8 | 93.0 | **0.39** | **6.9** | **116.1** |

### 2) *Comparison with the State-of-the-Art*

In order to comprehensively evaluate the performance of the DABNet-Lite in our experiment, it is compared with the state-of-the-art semantic segmentation models in both accuracy and speed. These models were trained in the same way as the DABNet-Lite on CTEM dataset. Table V shows the performance of various models in terms of prediction accuracy (mIOU), model parameters, float point of operations (FLOPs) and the prediction speed (fps). If images of 540×960 resolution are inputed in the network, the inference speed of DABNet-Lite on GTX1080 is 116 frames per second (fps) with 93.0% mIOU, showing great results on real-time performance and accuracy. A visualized comprehensive comparison of various models on performance is further demonstrated in Fig. 10, where the DABNet-Lite is on the top right. This model lays a soild foundation for model deployment on mobile devices.

TABLE V
COMPARISON WITH THE STATE-OF-THE-ART

| Method | mIOU(%) | Parameters(M) | FLOPs(G) | FPS |
|---|---|---|---|---|
| ESNet [50] | 90.1 | 1.66 | 24.12 | 41.7 |
| ESPNetV2 [51] | 92.6 | 1.25 | 5.64 | 61.3 |
| Fast-SCNN [52] | 89.5 | 1.14 | **1.74** | **209.6** |
| LEDNet [53] | 91.4 | 0.92 | 11.37 | 45.2 |
| ContextNet [54] | 90.7 | 0.87 | 1.75 | 186.6 |
| EDANet [55] | 90.5 | 0.68 | 8.84 | 82.1 |
| CGNet [56] | 92.8 | 0.49 | 6.90 | 55.4 |
| ENet [57] | 90.5 | **0.36** | 3.94 | 67.7 |
| DABNet [42] | **93.2** | 0.75 | 10.36 | 78.9 |
| DABNet-Lite | 93.0 | 0.39 | 6.91 | 116.1 |

### 3) *Trade-off Between Full Annotation and Partial Annotation*

In the training set, both fully annotated images and partially annotated images are included. Compared with the full labeling method, the partial labeling one is more economical in time and labor costs, thus it is also worth considering whether partial annotation can be adopted instead of fully annotation or how to balance the proportion of samples annotated in these two methods, especially when the images of the dataset is in a huge amount.

We fixed the number of images for training to 500 and changed the proportion of fully annotated images involved in. Fig. 11 presents the training results on samples in different annotations. When only partially annotated images were used in training, the mIOU on test set turned out above 88%, indicating that learning terrain properties with imprecise data can also achieve great results in this task. As the proportion of fully annotated images increases from 0% to 10%, the accuracy improved obviously about 4%. It is inferred that elaborate annotations offer more subtle details to learn fine-grained features for differentiation especially on terrain junction. In the process that the ratio of fully annotated images adds up to 20%, the increase of mIOU slows down. When the proportion further comes to 100% approximately, the precision only adds to nearly 1%, while the cost of annotation doubles or triples. In this semantic segmetantion task oriented to terrains, 20% is regared as a turning point where a small proportion of images in full annotation with large scale of rough labeled training samples used in training can also achieve competitive results with less labor cost.

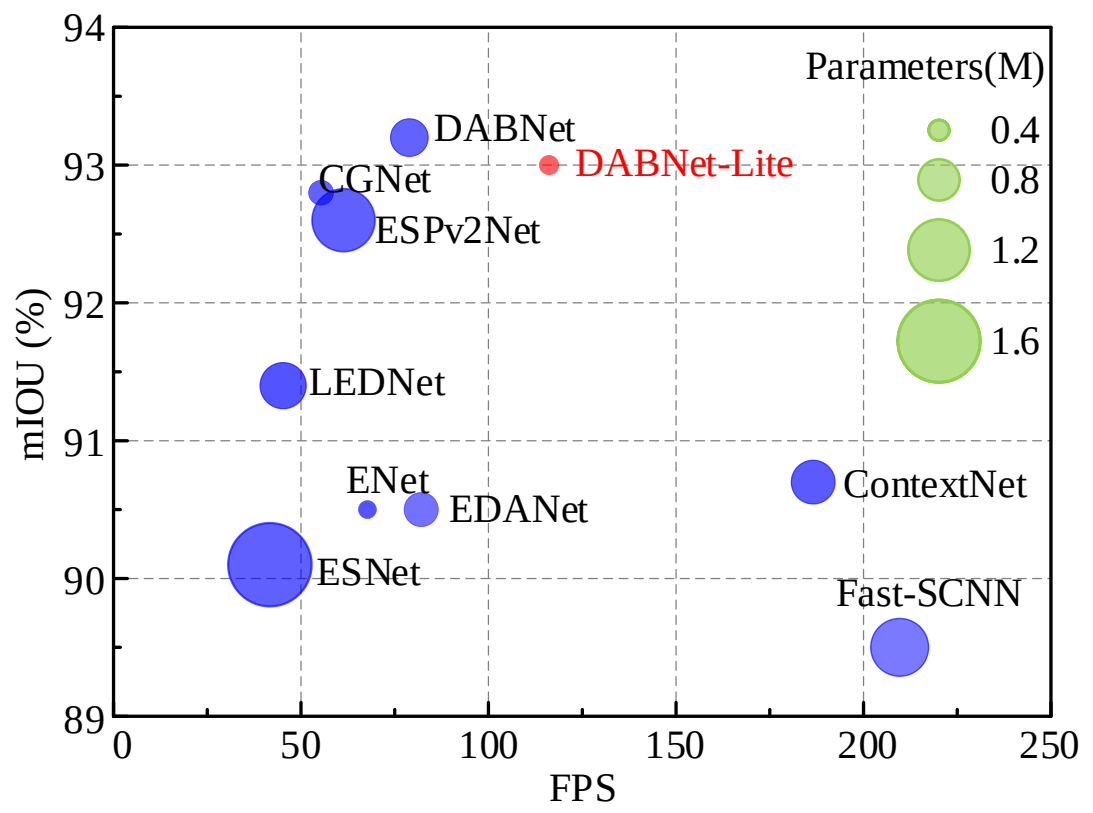

Fig. 10. Comparison results of various semantic segmentation models.

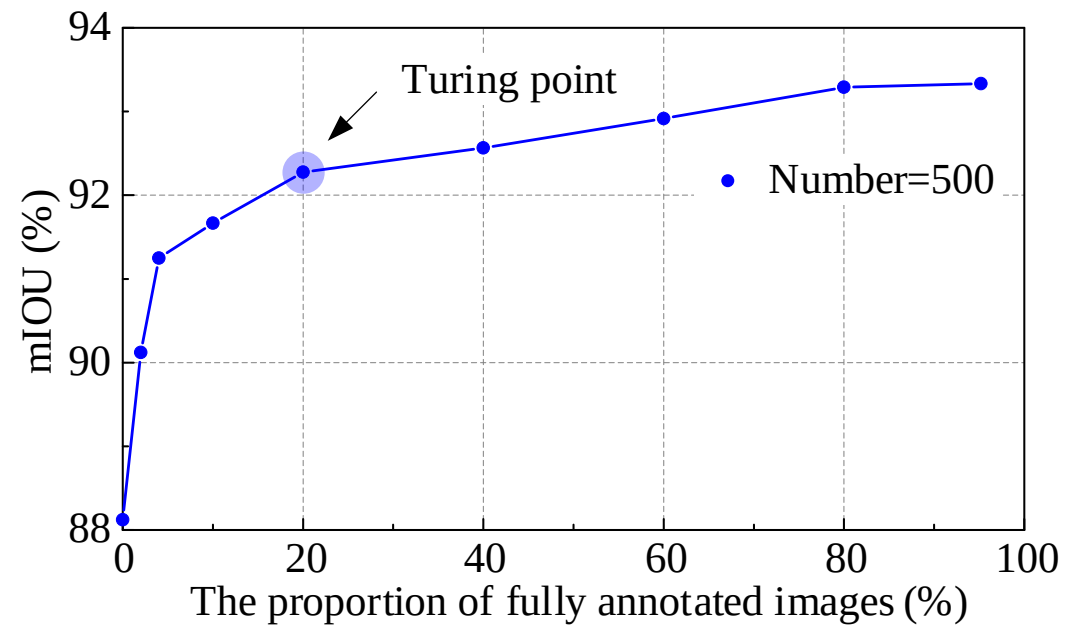

Fig. 11. The influence of the ratio of two annotation methods on the model training results.

### *B. Regression Results of Terrain Mechanical Properties Models*

The wheel force and torque data, the sinkage and the slip ratio data obtained in the data collection stage were filtered properly and down-sampled to the same frequency. By using the analytic formula of dominant terrain mechanical parameters mentioned in the third part, the identification results of various terrain mechanical properties obtained and corresponding regression model are shown in Fig. 12. It can be seen that the identification results of the dominant parameters of various terrains are almost in normal distribution. Soil and stony soil have similar mean value on sinkage exponent, and that for bedrock is a very small, approximate to 0.1, which means great stiffness. For internal friction angle $\varphi$, the identification results of bedrock are in large variance. This phenomenon attributes to the drastic vibration of rover when its lugged wheels contact with bedrock surface causing a large fluctuation on the forward speed. In turn, the slip ratio of the wheel which relates to the internal friction angle varies largely, thus there is a large variance in the identification result of bedrock. Such vibration also exists when driving on other terrain, resulting in a certain variance of the results, but it is drastic most on gravel and bedrock.

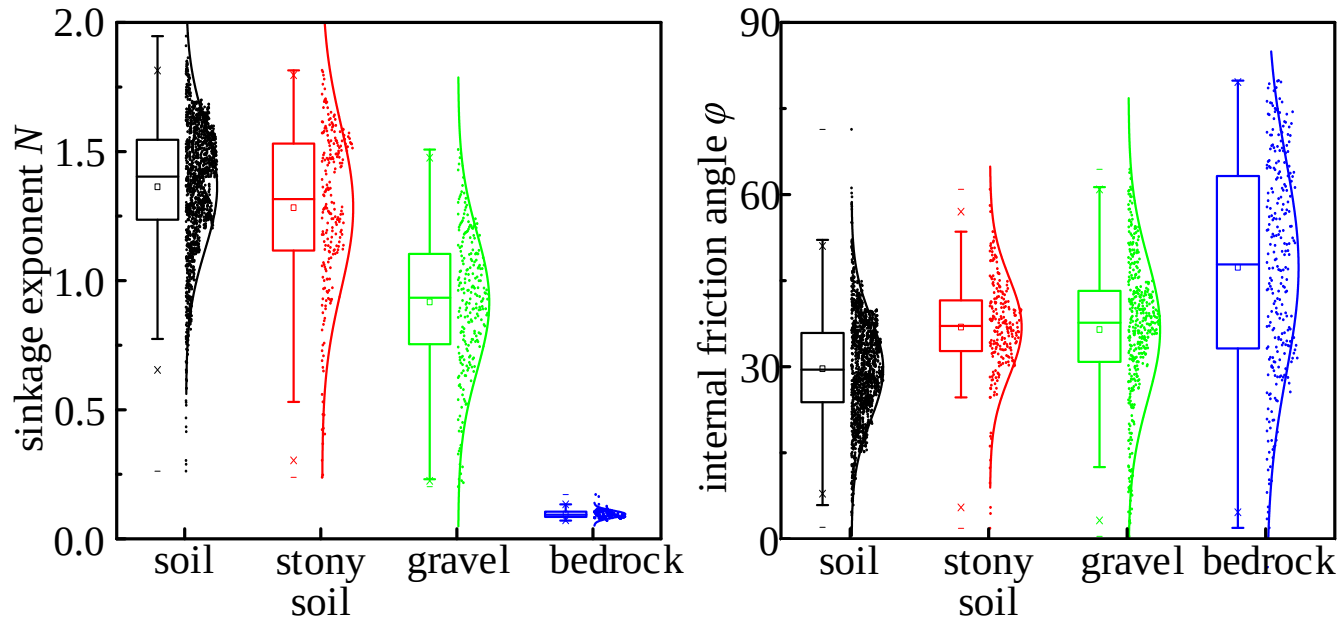

Fig. 12. Regression results of terrain mechanical properties models.

Specifically, the distribution of bearing and shearing properties for different terrains estimated by MLE algorithm are listed in Table VI. As the rock cannot be traversed to obtain the corresponding interactive data, the mechanical properties of rocks or bedrock are set with the maximum based on common sense. In addition, the mechanical properties of the background are not normally involved in planetary exploration missions, so all parameters are set to 0.

TABLE VI
REGRESSION PARAMETERS OF TERRAIN MECHANICAL PROPERTIES MODELS

| Parameters | soil | stony soil | gravel | bedrock | rock | background |
|---|---|---|---|---|---|---|
| $\mu(N)$ | 1.36 | 1.28 | 0.92 | 0.10 | 0.10 | 0 |
| $\sigma(N)$ | 0.25 | 0.32 | 0.27 | 0.01 | 0.01 | 0 |
| $\mu(\varphi)$ | 29.6 | 36.9 | 36.5 | 47.3 | 47.3 | 0 |
| $\sigma(\varphi)$ | 8.9 | 8.6 | 12.4 | 18.7 | 18.7 | 0 |

### *C. Results of Terrain Mechanical Properties Inference*

Combining the semantic segmentation results and the regressed bearing and shearing properties distribution model of various terrains, the predicted feature maps of the bearing and shearing properties are obtained. As displayed in Fig. 13, the bearing and shearing properties of different terrains are quite distinct, and there is a gentle transition at the junction of terrains, which is consistent with the subjective cognition.

The whole method of terrain property estimation is tested quantitatively in several scenarios. We compared the terrain bearing and shearing properties prediction results with ground truth gathered in the traversed routes. 11 routes are tested and one of the comparisons is provided in Fig. 14. It indicates that the bearing property prediction are consistent with the in-situ estimation results and most of the values are within the confidence bound. The in-situ estimation of the shearing property varied drastically due to the uneven surface of the gravel which cause rapid change in postures. The bearing property predicted from sight in the gravel phase has a significant gap with the in-situ estimation results, which results from the hard bedrock beneath the gravel. The average full-scale errors of the equivalent bearing and shearing properties estimated on all tested routes are 12.5% and 10.8%, and the probabilities that the ground truth are within the estimated interval is 0.585 and 0.623, respectively. It reveals that though it cannot get results accurate enough but most results are in the estimation bound which is helpful for path planning considering uncertainty but not appropriate for control.

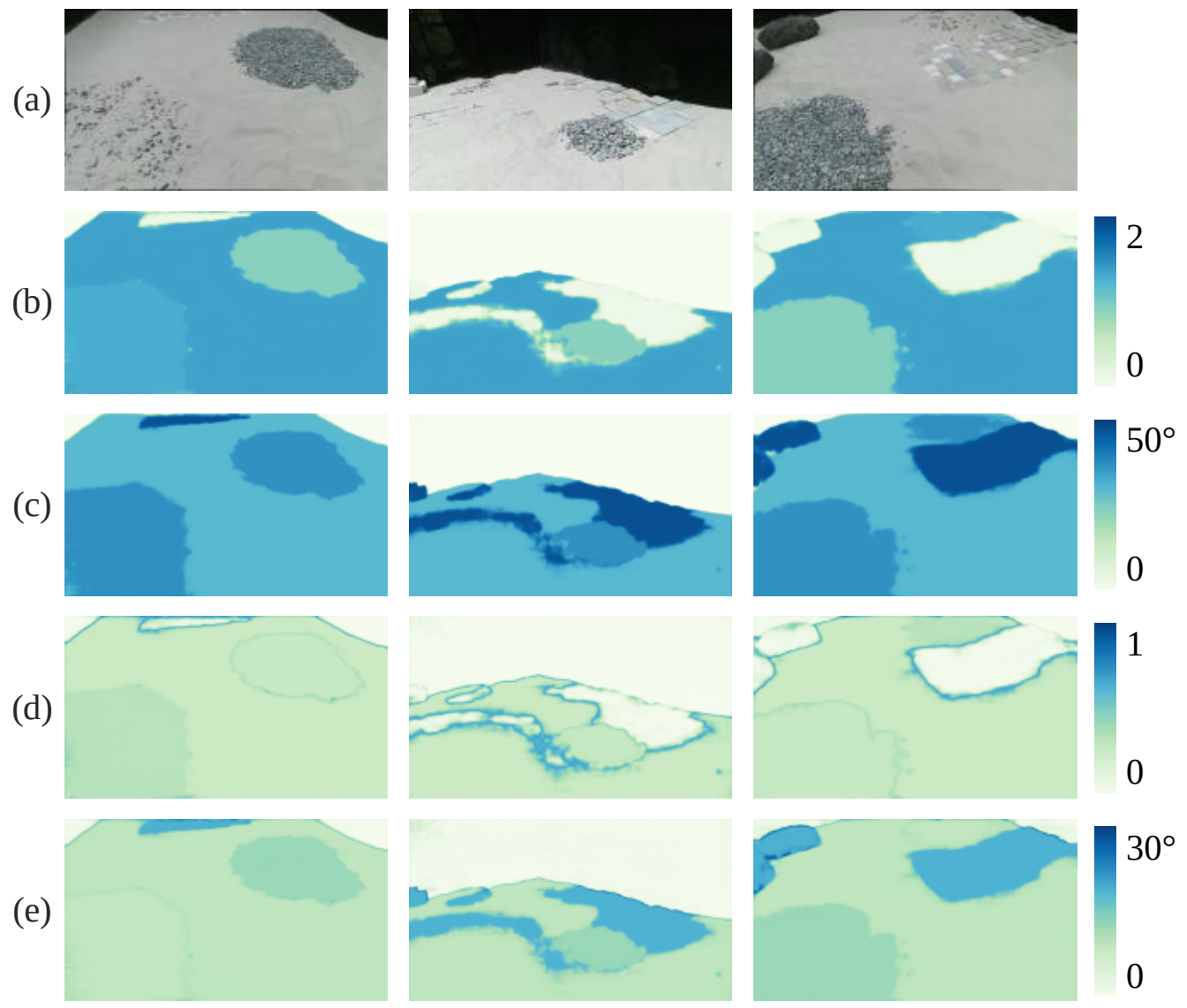

Fig. 13. Dense prediction results of terrain mechanical properties in full field of view. (a) original image; (b) mean value of sinkage exponent prediction; (c) mean value of internal friction degree prediction; (d) standard deviation of sinkage exponent prediction; (e) standard deviation of internal friction degree prediction.

In Figure 14, the predicted results are close to the in-situ estimations, and most of them fall within the confidence bound. However, there are fluctuations and a certainty degree of delay in the in-situ estimations at terrain junctions, such as when the rover moves from bedrock to sand as shown in the figure. Such phenomenon attributes to many reasons, but the most accountable one is that the in-situ estimation of dominant parameters with the aforementioned model is on the premise of the steady state of the wheel. At terrain junction, the motion state of the vehicle changes, and there will be some deviations

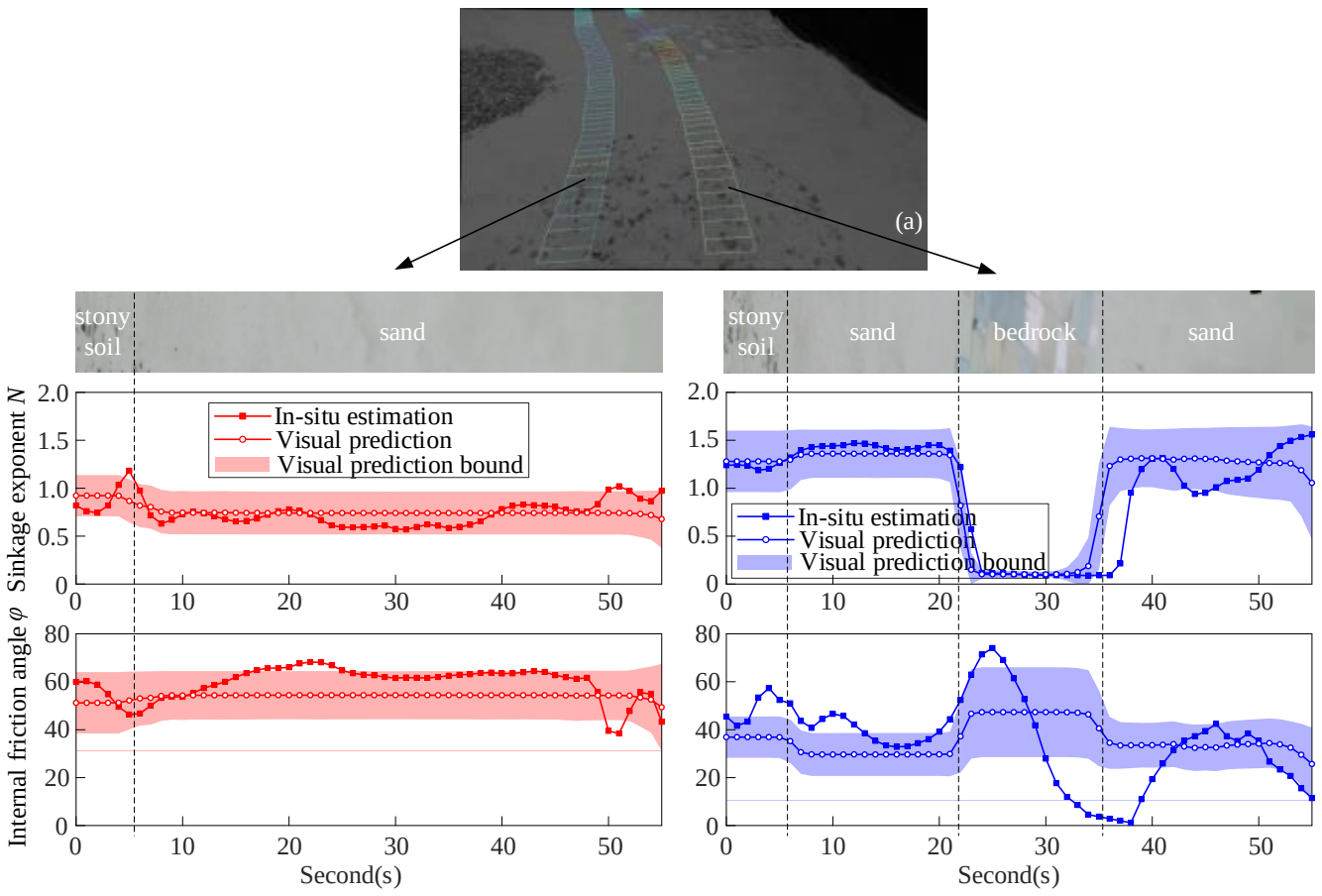

Fig. 14. Prediction results of dominant parameters of terrain mechanics for a path. (a) the image; (b) bearing property prediction of the left wheel; (c) bearing property prediction of the right wheel; (d) shearing property prediction of the left wheel; (e) shearing property prediction of the right wheel.

between the parameters identified by the unsteady data and the values predicted by vision.

In actual practice, the mechanical properties of a path planned before traverse can be predicted as shown in Fig. 15. The transition on terrain mechanical properties represented in the sinkage exponent and internal friction angle in terms of direction (getting bigger or smaller) and degree (how big the transition is) is predicted in advance which set aside time for adaptive locomotion mode switching or path re-planning. As the Fig. 15 illustrated, from the gravel to soil, there is a sharp increase in sinkage exponent and a mild decrease in shearing property which hits the rover to change from the wheeled mode to legged mode or follow the path with slip compensation. After traversed soil, the prediction shows that the upcoming bedrock in great stiffness and friction can support great driving force for moving forward. Before entering the stony soil where the stiffness is in great uncertainty, the rover can make active inspection tentatively for security.

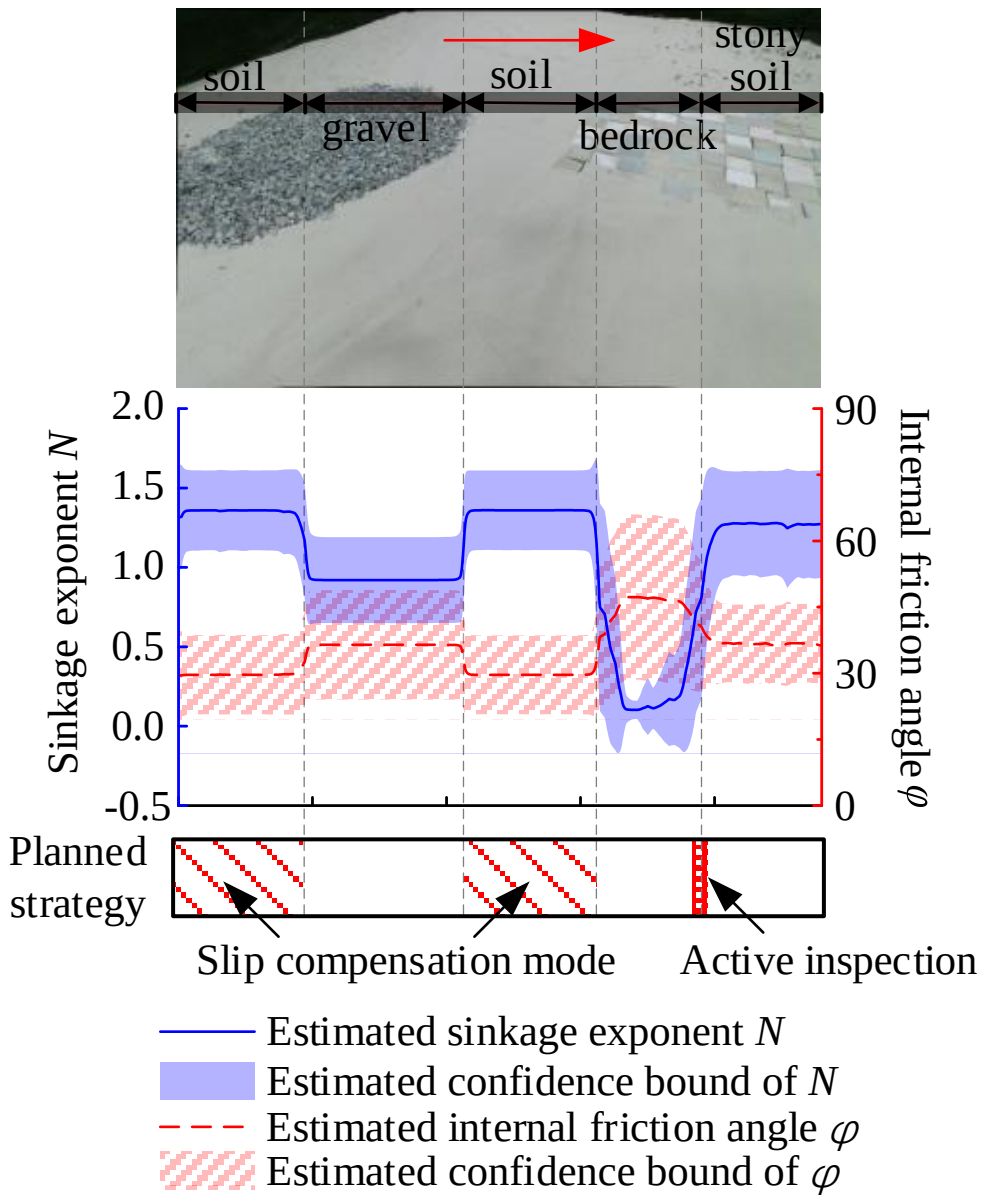

Fig. 15. Prediction results of dominant parameters of terrain mechanics for a path in terrain image.

## VI. Conclusion

In this paper, we presented a visual-based approach to predict terrain dominant mechanical properties including bearing and shearing properties in the distance. A two-staged prediction method linked with semantic clues is constructed which firstly classified planetary terrain in a pixel-wise level then transformed probabilistic semantic results to terrain properties values with empirical property regression models of each terrain. The method expands prediction results of terrain mechanical properties to a wider range and get more fine-grained results.

A semantic segmentation dataset of planetary terrains is established with two kinds of labels, making up the blank of research platform for semantic segmentation on planetary environment. A corresponding light-weighted terrain semantic segmentation model is realized with a competitive mIoU of 93.0% outperforming most of the networks on the established dataset. The compact model is only 0.39MB and achieves real-time computation of 116pfs on GTX1080 which is promising to be applied onboard. It also demonstrates that 20% training samples substituted with weakly labelled ones can save labelling efforts while without loss too much accuracy in this task. The whole method of terrain property estimation is quite accurate in terms of shearing but slightly inferior in bearing property and it helps to avoid non-geometric hazard in the planning stage.

Though our method predicts terrain categories and corresponding mechanical properties with satisfactory accuracy, it cannot deal with some deceptive scenarios, e.g. unconsolidated terrain hidden below duricrusts or dust, which depends more on rapid tactile perception and stuck recovery strategies.

As future work, a self-supervised method can be introduced to get rid of artificial labelling and promote the planetary exploration system to master environment mechanical properties in an autonomous manner. If more realistic terrain semantic segmentation dataset of planetary environment is available, the proposed model could be finetuned to adapt to the real planetary environment. In addition, the proposed terrain mechanical properties prediction method can also be useful for legged robots walking more adaptively on the wild.

## Acknowledgment

The authors would like to thank Lan Huang, Huanan Qi, in particular Lan Huang for developing and providing the mobile rover for experiments. They would like to thank Zhengyin Wang for his advice in terramechanics and Wenhao Lian for his contributions to the sensor system.